%% file: baeta.tex
\documentclass[runningheads]{llncs}

\usepackage{graphicx}
\usepackage{xcolor}
\usepackage{subcaption}

\usepackage{booktabs}
\usepackage{multirow}
\usepackage{censor}
\usepackage{xspace}
\usepackage{array, multirow}
\usepackage{makecell}
\usepackage{float}
\usepackage{dirtytalk}
\usepackage{tablefootnote}
\usepackage{mathtools}
\usepackage{cite}
\usepackage{longtable}

\usepackage{tabularx}
\newcolumntype{Y}{>{\centering\arraybackslash}X}

\usepackage{amsmath}
\usepackage{hyperref}
\DeclarePairedDelimiter\floor{\lfloor}{\rfloor}

\newcommand{\eg}{\textit{e.\,g.}\xspace}

\begin{document}

\title{TensorGP -- Genetic Programming Engine in TensorFlow}

\author{Francisco Baeta \and João Correia \and Tiago Martins \and Penousal Machado}
%\author{unknown}

\institute{CISUC, Department of Informatics Engineering, \\
University of Coimbra, Coimbra, Portugal \\
\email{\{fjrbaeta,jncor,tiagofm,machado\}@dei.uc.pt}}

\authorrunning{F. Baeta et al.}
%\authorrunning{unknown}

%\pagestyle{empty}

\maketitle

\begin{abstract}

In this paper, we resort to the TensorFlow framework to investigate the benefits of applying data vectorization and fitness caching methods to domain evaluation in Genetic Programming.
For this purpose, an independent engine was developed, TensorGP, along with a testing suite to extract comparative timing results across different architectures and amongst both iterative and vectorized approaches. Our performance benchmarks demonstrate that by exploiting the TensorFlow eager execution model, performance gains of up to two orders of magnitude can be achieved on a parallel approach running on dedicated hardware when compared to a standard iterative approach.


\end{abstract}

%\captionsetup[figure]{font=small,skip=0pt}

\input{1-introduction}

\input{2-relatedwork}

\input{3-tensorgp}

\input{4-experimentation}
\input{5-conclusion}

\bibliographystyle{splncs03}
\bibliography{baeta}

\end{document}

%% file: 1-introduction.tex
\section{Introduction}
\label{sec:intro}

Genetic Programming (GP), which targets the evolution of computer programs, is known to require large amounts of computational resources since all individuals in the population need to be executed and tested against the objective. As a result, fitness evaluation is generally regarded as the most computationally costly operation in GP for most practical applications \cite{9}.
Despite this, GP is beyond doubt a powerful evolutionary technique, capable of tackling every problem solvable by a computer program without the need for domain-specific knowledge \cite{poli2008field}. Furthermore, although computationally intensive by nature, GP is also \say{embarrassingly parallel} \cite{10}.

Previous works on accelerating fitness evaluation in GP mainly focus on two techniques: the caching of intermediate fitness results and the vectorization of the evaluation domain.
The first method aims to save the results of code execution from parts of a program to avoid re-executing this code when evaluating other individuals.
On the other hand, the second method evaluates the full array of fitness cases simultaneously by performing a tensor operation for each function within an individual.

The last decade saw the exponential growth of computing power proposed by Gordon Moore back in 1965 \cite{16} start to break down. As we start meeting the limits of physics, a paradigm shift towards multi-core computing and parallelization becomes inevitable.
Namely, with the rise of parallel computing, devices such as the Graphics Processing Units (GPUs) have become ever more readily available \cite{arenas2012gpu}.
Tensor operations are highly optimised on GPUs as they are necessary for the various stages of the graphical rendering pipeline. Therefore, it makes sense to couple the data vectorization approach with such architectures.

In this work, we resort to the TensorFlow plataform in order to investigate the benefits of applying the aforementioned approaches to the fitness evaluation phase in GP, as well as comparing performance results across different types of processors.
With this purpose, a novel and independent GP engine was developed: TensorGP.
Other engines such as KarooGP \cite{24} already take advantage of TensorFlow's capabilities to speed program execution. However, our engine exploits new TensorFlow execution models, which are shown to benefit the evolutionary process.
Moreover, we intend on extending the application of TensorGP outside the realm of classical symbolic regression and classification problems by providing support for different types of functions, including image specific operators.

The remainder of this paper is organised as follows. Section \ref{sec:relatedWork} provides a compilation of related work. Section \ref{sec:framework} presents the framework. Section \ref{sec:expsetup} lays the experimental setup and analyzes benchmarking results. Finally, Section \ref{sec:conclusions} draws final conclusions and points towards future work.

%% file: 2-relatedwork.tex
\section{Related Work} \label{sec:relatedWork}

Because GP individuals usually share highly fit code with the rest of the population and not only within themselves \cite{5}, techniques to efficiently save and reuse the evaluation of such code have been of special interest to research around GP.
In specific, Handley \cite{4} first implemented a method of fitness caching by saving the computed value by each subtree for each fitness case.
Furthermore, Handley represented the population of parsed trees as a Directed Acyclic Graph (DAG) rather than a collection of separate trees, consequently saving memory by not duplicating structurally identical subtrees.

However, because system memory is finite, the caching of intermediate results must obey certain memory constraints.
In this regard, Keijzer \cite{7} proposed two cache update and flush methods to deal with fixed size subtree caching: a first method using a postfix traversal of the tree to scan for nodes to be added to or deleted from the cache and a second method that implemented a variant of the DAG approach. 
Even if we rule out the amount of memory used, hit-rates and search times are still a grave concern.
Wong and Zhang \cite{8} developed a caching mechanism based on hash tables to estimate algebraic equivalence between subtrees, which proved efficient in reducing the time taken to search for common code by reducing the number of node evaluations. Besides, caching methods are particularly useful in scenarios with larger evaluation domains and where code re-execution is more time-consuming. As an example, Machado and Cardoso \cite{machado1999speeding} applied caching to the evolution of large-sized images (up to 512 by 512 pixels) in the NEvAr evolutionary art tool.

Another common way to accelerate GP is to take advantage of its potential for parallelization.
Various works have explored the application of parallel hardware such as Central Processing Units (CPUs) with Single Instruction Multiple Data (SIMD) capabilities \cite{26, 111, de2020mimd}, GPU-based architectures \cite{19, 14, 13, 20} and even Field Programmable Gate Arrays (FPGAs) \cite{25} to fitness evaluation within the scope of GP. 
However, arguably the most promising speedups still come from GPUs as they are the most widely available throughput-oriented architectures.
Namely, Cano et al. \cite{19} verified speedups of up to 820 fold for certain classification problems versus a standard iterative approach by massively parallelizing the evaluation of individuals using the NVIDIA Compute Unified Device Architecture (CUDA) programming model.

One common way to abstract this parallelization process is to vectorize the set of operations performed over the fitness domain, effectively reducing the running time of a program to the number of nodes it contains \cite{7}.
Some interpreted languages such as Matlab, Python and Perl already support vectorized operations in an attempt to reduce computational efforts. In particular, TensorFlow \cite{tensorflow} is a numerical computation library written in Python that provides an abstraction layer to the integration of this vectorization process across different hardware.
Staats et al. \cite{24} demonstrated the benefits of using TensorFlow to vectorized GP fitness data in both CPU and GPU architectures, achieving performance increases of up to 875 fold for certain classification problems.
The engine that the authors developed, KarooGP, is still used to tackle many symbolic regression and classification problems \cite{kai1, kai2, kai3}. However, KarooGP does not take advantage of recent additions to TensorFlow execution models.

%% file: 3-tensorgp.tex
\section{TensorGP} \label{sec:framework}

TensorGP takes the classical approach of most other GP applications and expands on it by using TensorFlow to vectorize operations, consequently speeding up the domain evaluation process through the use of parallel hardware. 
Moreover, TensorFlow allows for the caching of intermediate fitness results, which accelerates the evolutionary process by avoiding the re-execution of highly fit code. 
TensorGP is implemented in Python 3.7 using the TensorFlow 2.1 framework and is publicly available on GitHub \footnotemark[1].

%\footnotetext[1]{Repository available at: https://github.com/$<$ommited for blind review$>$}
\footnotetext[1]{TensorGP repository available at \url{https://github.com/AwardOfSky/TensorGP}}

In this section, we describe the implementation details of the incorporated GP features, as well as the efforts of integrating some of these features with the TensorFlow platform.

\subsection{Genotype to Phenotype} \label{sec:genotypetophenotype}

\begin{figure}[h]
  \centering
  \includegraphics[width=\linewidth]{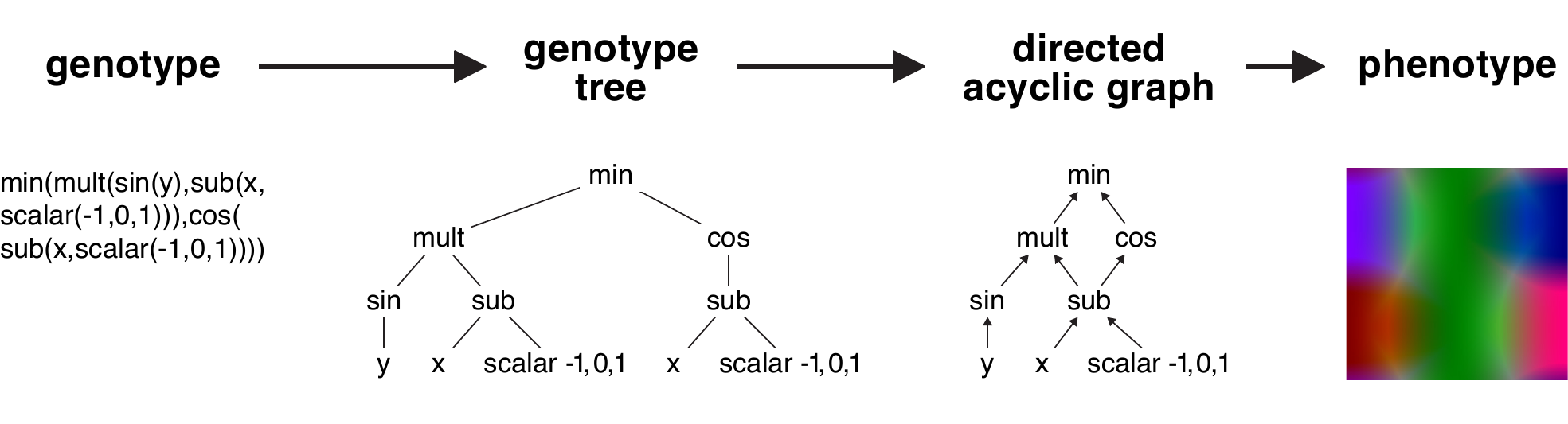}
  \caption{Genotype to phenotype translation phases in TensorGP.}
  \label{fig:genfen}
\end{figure}

As the name implies, TensorGP works with tensors. In essence, a tensor is a generalization of scalars (that have no indices), vectors (that have exactly one index), and matrices (that have exactly two indices) to an arbitrary number of indices \cite{wolfram}.
We start by describing the process of executing an individual in TensorGP.
Figure \ref{fig:genfen} demonstrates our engine's translation pipeline from genotype to phenotype.

In its simplest form, each individual in GP can be represented as a mathematical expression. TensorGP follows a tree-based approach, internally representing individuals as a tree graph. This implies a first translation phase from string to tree representation, which is only performed at the beginning of the evolutionary process in case the initial population is not randomly generated.

TensorFlow can either execute in an eager or graph-oriented mode. When it comes to graph execution, TensorFlow internally converts the tree structure into a graph before actually calculating any values. This is done in order to cache potential intermediate results from subtrees, effectively generalizing our tree graph structure to a DAG. On the other hand, the eager execution model allows for the immediate execution of vectorized operations, eliminating the overhead of explicitly generating the intermediate DAG of operations.

Even though graph-oriented execution enables many memory and speed optimizations, there are heavy performance costs associated with graph building.
TensorFlow eager execution mode aims to eliminate such overheads without sacrificing the benefits furnished by graphs \cite{agrawal2019tensorflow}.
Because the individuals we are evolving are constantly changing from generation to generation, we would be inclined to think that eager mode would be a good fit for tensor execution. For this reason, in Section \ref{sec:expsetup}, we include some performance comparisons between both these TensorFlow execution modes.

Finally, the last translation phase goes through the entire genotype data to produce a phenotype, which will be the target of fitness assessment.
Because the domain of fitness data points to be evaluated is fixed for all operations, the vectorization of this data is made trivial using a tensor representation.
Generally speaking, our phenotype is a tensor, which can be visually represented as an image for a problem with 3 dimensions, as seen in the last stage of Figure \ref{fig:genfen}. In this example, the first two dimensions correspond to the width and height of the image, while the third dimension encodes information regarding the RGB color channels.
The resulting tensor phenotype is obtained by chaining operators, variables and constants that make part of the individual. These variables and constants are also tensors, which occupy a range of $[-1, 1]$ for the example given.
With the aid of TensorFlow primitives, we can apply an operation to all domain points at the same time while seamlessly distributing computational efforts amongst available hardware.

\subsection{Primitive Set} \label{sec:primitiveset}

To provide a general-purpose GP tool and ease evolution towards more complex solutions, the primitive set implemented goes beyond the scope of simple mathematical and logic operators. This way, we attempt to provide sufficiency through redundancy of operators for as many problems as possible. Some image specific operators are also included to facilitate the application of TensorGP to image evolution domains (such as evolutionary art).
One of such operators, and perhaps the most intriguing, is the $warp$.
The $warp$ operator is commonly used to deform images and is defined as a transformation that maps every element of a tensor to a different coordinate. This mapping is done according to an array of tensors with a size equal to the number of dimensions of the problem. Each of these tensors dictate where elements will end up within a given dimension.

Operators must also be defined for all possible domain values, which in some cases means implementing protection mechanisms for certain cases. Table \ref{tab:operator_desc} enumerates the different types of operators and respective special cases. All operators are applied to tensors and integrated into TensorGP through composition of existing TensorFlow functions.
While the main math and logic operators can be implemented with a simple call to the corresponding TensorFlow function, more complex operators may imply chaining multiple functions.
For instance, while TensorFlow possesses a plethora of operators to cater to our vectorization needs, it lacks a $warp$-like operator. Therefore, to implement it, we need to express the whole transformation process as a composition of existing TensorFlow functions.

Besides the specified protective mechanism, there are noteworthy implementation details for some operators.
As an example, when calculating trigonometric operators, the input argument is first multiplied by $\pi$. The reasoning behind this is that most problem domains are not defined in the  [$-\pi$, $\pi$] range, but are otherwise normalised to either [0, 1] or [-1, 1]. As a matter of fact, [-1, 1] this is the standard domain range used in TensorGP. This makes it so that the argument to the trigonometric operators is in the [$-\pi$, $\pi$] range, which covers the whole output domain for these operators.

\begin{table}[]
\caption{Description of implemented TensorGP operators.}
    \centering
\begin{tabular}{|*{5}{c|}}
    \hline
 \textbf{Type}    &    \textbf{Subtype} &  \makecell{\textbf{Operator} \\ (engine abbreviation)}  & \textbf{Arity}  & \textbf{Functionality} \\
    \hline
\multirow[t]{19}{*}{Mathematical}           
            & \multirow[t]{4}{*}{Arithmetic}
                        & Addition (add)            & 2 &  $x + y$                            \\  \cline{3-5}
            &           & Subtraction (sub)         & 2 &  $x - y$                          \\  \cline{3-5}
            &           & Multiplication (mult)     & 2 &  $x \times y$                            \\  \cline{3-5}
            &           & Division (div)            & 2 &  \makecell{$x / y$ \\ 0 if denominator is 0}    \\  \cline{2-5}
            & \multirow[t]{3}{*}{Trigonometric $^a$}
                        & Sine (sin)                & 1 & \makecell{$\cos(x \pi)$}  \\  \cline{3-5}
            &           & Cosine (cos)              & 1 & \makecell{$\sin(x \pi)$}  \\  \cline{3-5}
            &           & Tangent (tan)             & 1 & \makecell{$\tan(x \pi)$}  \\  \cline{2-5}
            & \multirow[t]{12}{*}{Others}
                        & Exponentional (exp)       & 1 & \makecell{$e^{x}$}  \\  \cline{3-5}
            &           & Logarithm (log)           & 1 & \makecell{$\log{x}$ \\ -1 if $x < 0$}  \\  \cline{3-5}
            &           & \makecell{Exponentiation \\ (pow)}      & 2 & \makecell{$x ^ y$ \\ 0 if $x$ and $y$ equal  0  }  \\ \cline{3-5}
            &           & Minimun (min)             & 2 & \makecell{$min(x, y)$}  \\ \cline{3-5}
            &           & Maximun (max)             & 2 & \makecell{$max(x, y)$}  \\ \cline{3-5}
            &           & Average (mdist)           & 2 & $(x + y) / 2$  \\ \cline{3-5}
            &           & Negative (neg)            & 1 & \makecell{-$x$}  \\ \cline{3-5}
            &           & \makecell{Square Root \\ (sqrt)}  & 2 & \makecell{$\sqrt{x}$ \\ 0 if $x < 0$}  \\ \cline{3-5}
            &           & Sign (sign)               & 1 & \makecell{-1 if $x < 0$ \\ 0 if $x$ equals 0 \\ 1 if $x > 0$}  \\ \cline{3-5}
            &           & \makecell{Absolute value \\ (abs)}      & 1 & \makecell{$-x$ if $x < 0$ \\ $x$ if $x \geq 0$}  \\ \cline{3-5}
            &           & Constrain (clip)$^b$      & 3 & \makecell{ensure $y \leq x \leq z$ \\ or $max(min(z, x), y)$ }  \\ \cline{3-5}
            &           & Modulo (mod)              & 2 & \makecell{$x \bmod y$ \\ remainder of division} \\ \cline{3-5}
            &           & \makecell{Fractional part$^b$ \\ (frac)}   & 1 & \makecell{\begin{math}x - \floor*{x}\end{math}} \\ 
    \hline
    \multirow[t]{6}{*}{Logic}           
            & \multirow[t]{4}{*}{Conditional}
                        & Condition (if)            & 3 & \makecell{if $x$ then $y$ else $z$}  \\  \cline{2-5}
            & \multirow[t]{3}{*}{Bitwise $^c$}
                        & OR (or)                   & 2 & \makecell{ logic value of $x \lor y$ \\ for all bits }  \\  \cline{3-5}
            &           & Exclusive OR (xor)        & 2 & \makecell{ logic value of $x \oplus y$ \\ for all bits}  \\  \cline{3-5}
            &           & AND (and)                 & 2 & \makecell{ logic value of $x \land y$ \\ for all bits}  \\
    \hline
    \multirow[t]{6}{*}{Image}           
            & \multirow[t]{1}{*}{Transform}
                        & Warp (warp)               & n & \makecell{ Transform data \\ given tensor input $^d$ }  \\  \cline{2-5}
            & \multirow[t]{3}{*}{Step}
                        & Normal (step)                         & 1 & \makecell{-1 if $x < 0$ \\ 1 if $x >= 0$}  \\  \cline{3-5}
            &           & Smooth (sstep)                        & 1 & \makecell{$x^2 (3 - 2x)$}  \\  \cline{3-5}
            &           & \makecell{Perlin Smooth \\ (sstepp)}  & 1 & \makecell{$x^3 (x (6x - 15) + 10)$}  \\ \cline{2-5}
            & \multirow[t]{2}{*}{Color}
                        & Distance (len)                            & 2 & \makecell{$\sqrt{x^2 + y^2}$}  \\  \cline{3-5}
            &           & \makecell{Linear Interpolation} (lerp)    & 3 & \makecell{$x + (y - x) \times frac(z)$}  \\
    \hline
\end{tabular}
\
\footnotesize{$^a$ Input argument in radians, $^b$ These are mostly support operators $^c$ Transformation to integer is needed, $^d$ More details in Subsection \ref{sec:primitiveset}}\\
%\footnotesize{$^d$ More details in Section \ref{sec:primitiveset}}\\
\label{tab:operator_desc}
\end{table}

\subsection{Features} \label{sec:Features}

TensorGP was implemented with ease of use in mind. To demonstrate some of its functionality, the following paragraphs describe the main features of the presented engine.

When a GP run is initiated on TensorGP, a folder is created in the local file system with the aim of logging evolution data.
In each generation, the engine keeps track of depth and fitness values for all individuals. 
When the run is over, a visualzation for the depth and fitness values of individuals across generations is automatically generated along with a CSV file with experimental data.

%\item[$\bullet$ Engine State]
Besides, TensorGP keeps an updated state with all the important parameters and evolution data. 
%The idea is to be able to pause the evolutionary process at any time and close the program to resume the experiment later. To do this, when the engine is called, it starts by creating a configuration file for that experiment with the appropriate settings (\eg population size, generations, mutation rate). 
With each new generation, the engine updates this file with information regarding evolution status. When it is time to resume the experiment, the engine simply loads the corresponding configurations from the file of that experiment, gathering the latest generational data.

%\item[$\bullet$ Custom Initial Population]
Although the default engine behavior is to generate the initial population according to a given (or otherwise random) seed, the user can choose to specify a custom initial population by passing a text file containing string-based programs to the engine. 

%\item[$\bullet$ Stop criteria]
Currently, there are two stop criteria implemented: the generation limit (which the engine defaults to) and acceptable error. In the acceptable error method, the experiment comes to an end if the best-fitted individual achieves a fitness value specified by the user. The conditional check for this value is made differently depending on whether we are dealing with a minimization or maximization problem, which leads to the next main feature.

It is possible to define custom operators for the engine.
The only requirement for the implementation of any operator is that it must returns a tensor generated with TensorFlow and have $dims=[ ]$ as one of the input arguments (in case the tensor dimensions are needed).
Along with the implementation, the user is required to register the operator by adding an entry to the function set with the corresponding operator name and arity.

%% file: 4-experimentation.tex
\section{Benchmark Experimentation} \label{sec:expsetup}

In this section, we describe the experimentation performed with the objective of investigating how TensorGP fares against other common GP approaches. To attain this goal, execution times were extracted and compared for a symbolic regression task across different domain sizes.
First, a tree evaluation experiment is implemented where we isolate the tensor evaluation phase and execute a batch of populations within a controlled environment. Next, we extend on this experiment by including the evolutionary process with the intent of testing a more typical GP scenario.

\subsection{Experimental Setup} \label{sec:experimentalset}

All experiments concern the symbolic regression problem of approximating the polynomial function defined by:

%\[\frac{1}{1+ x^{-4}} +  \frac{1}{1+ y^{-4}}\]
 \begin{equation}
 f(x,y)=\frac{1}{1+ x^{-4}} +  \frac{1}{1+ y^{-4}}
  \label{eq:pagie}
\end{equation}

This function is also known as Pagie Polynomial and is commonly used in GP benchmarks due to its reputation for being challenging \cite{pagie1997evolutionary}. Because the domain of this problem is two-dimensional, we represent it using a rank 2 tensor.

\begin{table}[hbt!]
\caption{\small Experimental GP parameters.} \label{tab:gpparams}
\centering
\begin{tabular}{l|l}
\hline
\multicolumn{1}{c|}{\textbf{Parameter}} & \multicolumn{1}{c}{\textbf{Value}}   \tabularnewline \hline
Runs     & 5 \tabularnewline
Maximum tree depth    & 12 \tabularnewline
Population size   & 50 \tabularnewline
Generation method   & Ramped-Half-and-Half \tabularnewline
Objective & Minimize RMSE from Pagie Polynomial \tabularnewline
Test cases & 6 ($2^{2n}$, for $n$ in $[6, 11]$)\tabularnewline
\hline
\end{tabular}
\end{table}

Table \ref{tab:gpparams} enumerates the parameters used for both the tree evaluation and evolutionary experiments. Regarding problem size, all experiments encompass the same array of 6 test cases, evaluating a two-dimensional domain that exponentially increases in size.
The first test case evaluates a 64 by 64 grid of data points, thus involving a total of 4,096 evaluations. In each subsequent test case, the length of the grid doubles, effectively quadrupling the number of points to evaluate. This grid of values, represented in TensorGP by a rank-2 tensor, keeps increasing until the length of each side of the tensor is 2,048 (over 4 million evaluation points).

Larger domains were not tested mainly due to VRAM limitations of the GPU used during the experiments.
Moreover, the same set of populations were used for all test cases, where each population contains 50 individuals generated with the Ramped-Half-and-Half method and 12 for maximum allowed depth.

While the first experiment only saw the execution of the aforementioned population batch, in the evolutionary run we let the individuals evolve for 50 generations for all test cases. Both experiments use the minimization of the Root Mean Squared Error (RMSE) metric as a fitness evaluation function.

In total, six approaches were considered. Four of these approaches concern TensorGP implementations, testing both graph and eager execution modes when running in the GPU versus CPU. 
The other two approaches implement serial GP evaluation methods: one resorting to the DEAP framework and another one using a modified version of the engine that evaluates individuals with the $eval$ Python function instead of TensorFlow.

DEAP is a commonly used EC framework written in Python and offers a powerful and simple interface for experimentation \cite{deap}.
We have chosen to include comparisons to this framework because it represents the standard for iterative domain evaluation in research and literature around GP. Moreover, DEAP is easy to install and allows for the prototyping of controlled environments within a few lines of code.

Furthermore, we also include our own serial baseline.
The purpose is to compare achievable timings for an iterative approach that does not use third-party software.
We do this by passing the expression of an individual to the Python $eval$ method so we can run it.
In order to eliminate the overhead of parsing code, $eval$ is called only once by plugging in a lambda defined expression of the individual.
For future reference, this approach will be referred to as EVAL.

\begin{table}[hbt!]
\caption{\small Hardware and software specifications used for all experiments.} \label{tab:testbed}
\centering
\begin{tabular}{l|l}
\hline
\multicolumn{1}{c|}{\textbf{Component}} & \multicolumn{1}{c}{\textbf{Specification}}    \tabularnewline \hline
CPU     & Intel® Core™ i7-6700 (3.40 GHz) \tabularnewline 
GPU    & NVIDIA GeForce GTX 1060 3GB   \tabularnewline
RAM   & 2 $\times$ 8GB @2,666 MHz \tabularnewline 
Operative System   & Windows 10  \tabularnewline 
Execution Environment & Command Prompt \tabularnewline 
\hline
\end{tabular}
\end{table}

The software and processing hardware used for the execution of these experiments is defined in Table \ref{tab:testbed}.

Because the experimental results presented in the next subsection encompass a wide range of values covering multiple orders of magnitude, our best bet for graphical representation is to use a logarithmic scale as it would be otherwise impossible to distinguish between timings.

\subsection{Results} \label{sec:evolrun}

\begin{figure}[hbt!]
    \centering
    \includegraphics[width=1.0\textwidth]{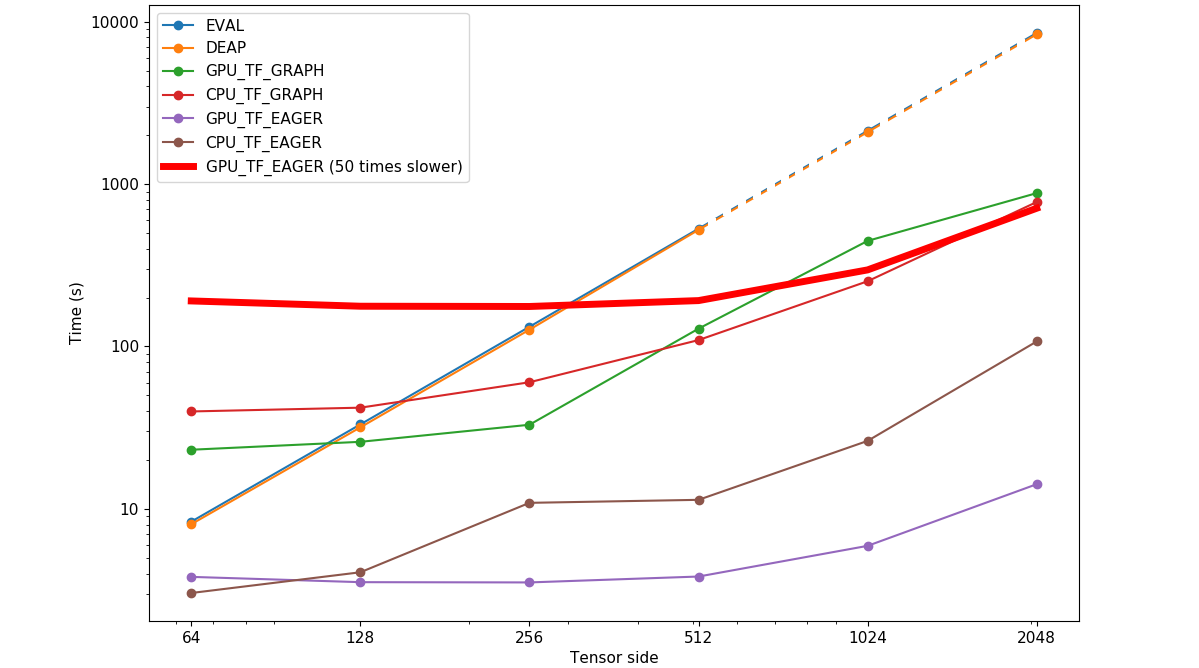}
    \caption{\small Time (in seconds) comparison of different approaches for raw tree evaluation across domain sizes.}
    \label{fig:graf1}
\end{figure}

The first experiment compares average execution times amongst all considered approaches.
Figure \ref{fig:graf1} shows the average time taken for the evaluation of all 5 populations in all test cases.
We can conclude that both EVAL and DEAP results are similar, following a linear increase in evaluation time with an increase in evaluation points.

\begin{table}[ht]
\caption{\small Standard deviation (STD) and average (AVG) of timing values (in seconds) across domain sizes for the tree evaluation experiment.}
    \centering
\begin{tabular}{ | c | c | r | r | r | r | r | r |p{0.7\linewidth}|}
    \cline{3-8}
 \multicolumn{2}{c|}{} & \textbf{EVAL} & \textbf{DEAP} & 
 \makecell{\textbf{TF graph} \\ \textbf{(GPU)}}  & \makecell{\textbf{TF graph} \\ \textbf{(CPU)}}  
 & \makecell{\textbf{TF eager} \\ \textbf{(GPU)}} &  \makecell{\textbf{TF eager} \\ \textbf{(CPU)}} \\
    \hline
\multirow[t]{2}{*}{}           
            $\mathbf{64 ^ 2}$ & \textbf{AVG} & 8.35 & 8.05 & 23.14 & 39.81 & 3.82 & 3.04 \\ %\cline{2-3}
            4,096 & \textbf{STD}  & 0.22 & 0.20 & 0.56	& 0.84 & 0.22 & 0.08 \\ \hline
\multirow[t]{2}{*}{}           
            $\mathbf{128 ^ 2}$ & \textbf{AVG} & 33.13 & 31.78 & 25.89 & 42.00 & 3.54 & 4.07 \\ %\cline{2-3}
            16,384 & \textbf{STD}  & 0.82 & 0.89 & 0.50	& 1.07 & 0.13 & 0.11 \\ \hline
\multirow[t]{2}{*}{}           
            $\mathbf{256 ^ 2}$ & \textbf{AVG} & 132.28 & 126.93 & 32.97 & 60.22 & 3.53 & 10.90 \\ %\cline{2-3}
            65,536 & \textbf{STD}  & 3.36 & 3.50 & 0.77	& 1.66 & 0.13 & 0.11 \\ \hline
\multirow[t]{2}{*}{}           
            $\mathbf{512 ^ 2}$ & \textbf{AVG} & 531.15 & 522.53 & 128.75 & 109.53 & 3.83 & 11.37 \\ %\cline{2-3}
            262,144 & \textbf{STD}  & 12.76 & 13.97 & 2.79	& 2.92 & 0.19 & 0.92 \\ \hline
\multirow[t]{2}{*}{}           
            $\mathbf{1,024 ^ 2}$ & \textbf{AVG} & DNF & DNF & 446.90 & 252.23 & 5.92 & 26.23 \\ %\cline{2-3}
            1,048,576 & \textbf{STD}  & DNF & DNF & 8.94	& 21.27 & 0.23 & 1.06 \\ \hline
\multirow[t]{2}{*}{}           
            $\mathbf{2,048 ^ 2}$ & \textbf{AVG} & DNF & DNF & 879.74 & 775.54 & 14.20 & 107.42 \\ %\cline{2-3}
            4,194,304 & \textbf{STD}  & DNF & DNF & 31.78 & 58.38 & 0.37 & 4.74 \\ \hline

\end{tabular}
\\
\footnotesize{\centering DNF stands for \say{Did Not Finish}.}
\label{tab:timingdordecabeca}
\end{table}

Because there is no domain vectorization, the direct relation between elements and time taken comes as no surprise.
It is worth noting that because of time constraints, results corresponding to the dashed lines in the two largest tensors sizes were not run but instead predicted by following the linear behavior from previous values.

Analysing results for TensorFlow eager execution mode, we see that even though the CPU is faster for the smallest test case, this trend fades rapidly for larger problems domains.
In fact, for 16,384 elements (\(128^2\)), the GPU is already marginally faster than the CPU. This margin widens with an increase in tensor side, resulting in GPU evaluation over a 4 million point domain (\(2,048 ^ 2\)) being almost 8 times faster as seen in Table  \ref{tab:timingdordecabeca}. As a reminder, with TensorFlow we are already providing operator vectorization, hence the 8 fold increase is
merely a product of running the same approach on dedicated hardware.

Moreover, we can also confirm the hypothesis that the evolutionary process in GP benefits from eager execution. For the two smaller domain sizes, we observed that the slowest eager
execution approach is about 10 times faster than the fastest graph execution one. This trend continues for larger domains but the gap shortens to about 8 times (in favor of eager execution).

Nevertheless, results gathered for graph execution show rather unexpected behavior. As suggested by the previously analyzed results, it would be safe to assume that the CPU would be faster for small domains with the GPU taking over for larger ones. In reality, the opposite is happening: the GPU is faster for domains up to 65,536 (\(256 ^ 2\)) in size, from which point on the CPU takes over. The answer to this strange behavior may lie in the graph implementation used.
Fitting every individual of a population in one session graph proved to take too much memory for larger domains, making these approaches even slower both in GPU and CPU.
This need for memory is specially taxing for the GPU VRAM (which is only 3GB compared to the available 16GB for system memory) that did not even finish some test cases while trying to include the entire population in a single graph.

\begin{figure}[h]
    \centering
    \includegraphics[width=1.0\textwidth]{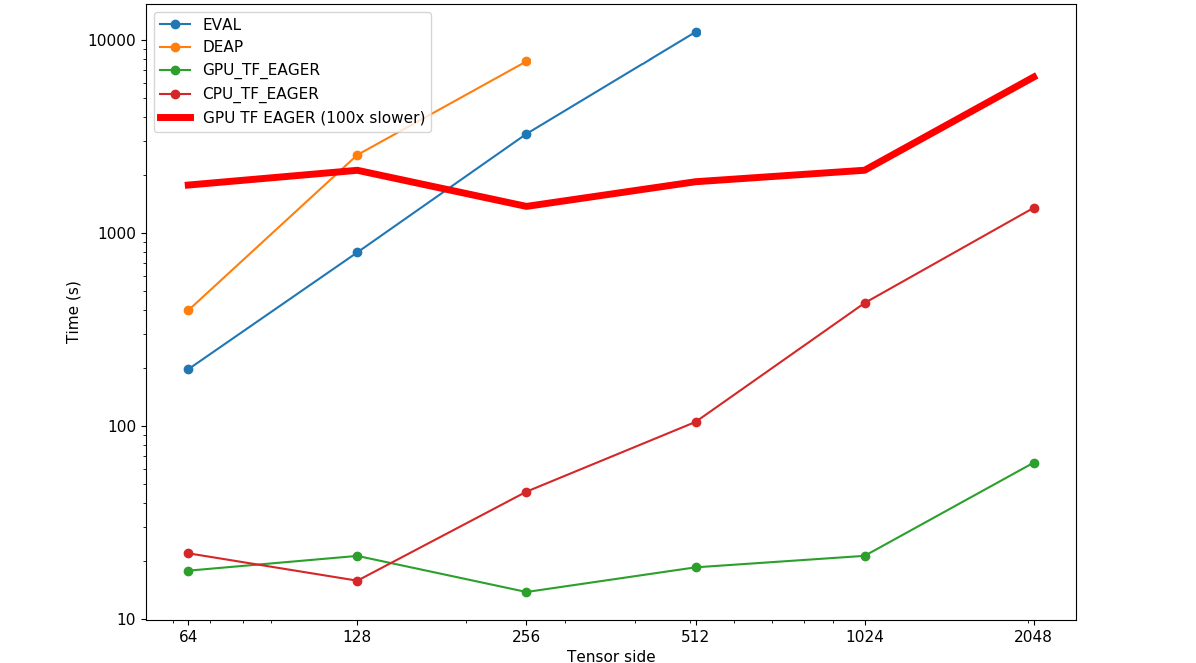}
    \caption{\small Time (in seconds) comparison of different approaches for a full evolutionary run across domain sizes.}
    \label{fig:mainfullrungraph}
\end{figure}

For this reason, and to be consistent with all domain sizes, we decided to test these graph execution approaches by opening a session graph for each individual instead of evaluating the entire population in a single graph.
Still, we can safely conclude that both graph execution approaches are slower than their eager equivalents.
The red bold line, in Figure \ref{fig:graf1}, is intended as a visualization aid that represents a 50 times performance threshold over our fastest approach (TensorFlow eager on GPU) for approaches above the line.

For the evolutionary experiment, however, only four of the considered approaches were used: the iterative ones (DEAP and EVAL) and the ones that concern TensorFlow execution in eager mode (both in CPU and GPU). Graph execution was omitted as it was demonstrated to be systematically slower than their eager equivalents.

\captionsetup[table]{font=small,skip=0pt}

\begin{table}[ht]
\caption{\small Standard deviation (STD) and average (AVG) timing values (in seconds) across domain sizes for the evolutionary experiment.}
    \centering
\begin{tabular}{ | c | c | r | r | r | r |}
    \cline{3-6}
 \multicolumn{2}{c|}{} & \makecell{\textbf{EVAL}} & \makecell{\textbf{DEAP}} 
 & \makecell{\textbf{TF eager} \\ \textbf{(CPU)}} &  \makecell{\textbf{TF eager} \\ \textbf{(GPU)}} \\
    \hline
\multirow[t]{2}{*}{}           
            $\mathbf{64 ^ 2}$ & \textbf{AVG} & 196.87 & 397.82 & 21.86 & 17.77 \\ %\cline{2-6}
            4,096 & \textbf{STD} & 96.26 & 339.50 & 13.44 & 8.29 \\ \hline
\multirow[t]{2}{*}{}           
            $\mathbf{128 ^ 2}$ & \textbf{AVG} & 795.36 & 2546.83 & 15.76 & 21.19 \\ %\cline{2-6}
            16,384 & \textbf{STD} & 381.71 & 2127.38 & 7.85 & 11.99 \\ \hline
\multirow[t]{2}{*}{}           
            $\mathbf{256 ^ 2}$ & \textbf{AVG} & 3274.13 & 7783.76 & 45.64 & 13.77 \\ %\cline{2-6}
            65,536 & \textbf{STD} & 2482.256 & 5824.78  & 20.79 & 7.39 \\ \hline
\multirow[t]{2}{*}{}           
            $\mathbf{512 ^ 2}$ & \textbf{AVG} & 11052.16 & DNF & 104.96 & 18.49 \\ %\cline{2-6}
            262,144 & \textbf{STD} & 3887.54 & DNF & 30.21 & 8.75 \\ \hline
\multirow[t]{2}{*}{}           
            $\mathbf{1,024 ^ 2}$ & \textbf{AVG} & DNF & DNF & 434.37 & 21.21 \\ %\cline{2-6}
            1,048,576 & \textbf{STD} & DNF & DNF & 160.05 & 9.98 \\ \hline
\multirow[t]{2}{*}{}           
            $\mathbf{2,048 ^ 2}$ & \textbf{AVG} & DNF & DNF & 1353.67 & 64.54 \\ %\cline{2-6}
            4,194,304 & \textbf{STD} & DNF & DNF & 679.56 & 32.51 \\ \hline
\end{tabular}
\\
\footnotesize{\centering DNF stands for \say{Did Not Finish}.}
\label{tab:stdfinalfullrun}
\end{table}

Figure \ref{fig:mainfullrungraph} shows the total run time for all considered approaches. Probably the most noticeable aspect of these results is that they appear to be less linear when compared to those regarding raw tree evaluation. This happens because, even though we are using a fix population batch for each test case, evolution might be guided towards different depths for different initial populations. If the best-fitted individual happens to have a lower depth value, the rest of the population will eventually lean towards that trend, lowering the overall average population depth and thus rendering the tensor evaluation phase less computationally expensive.
The opposite happens if the best-fitted individual is deeper, resulting in more computing time.
This explains the relatively higher standard deviations presented in Table \ref{tab:stdfinalfullrun} and the non-linear behavior across problem sizes (\eg the test case for size 65,536 (\(256 ^ 2\)) runs faster than the two smaller domains for TensorFlow running on GPU). 

Still regarding TensorFlow results, from the two first test cases, we can not identify a clear preference towards CPU or GPU as, for these domain sizes, the GPU memory transfer overhead is in par with the lack of CPU parallelization power. Nonetheless, for test cases larger than 65,536 (\(256 ^ 2\)), a clear preference towards GPU starts to be evident, with an average speedup of over 21 times for a problem with 4 over million points (\(2,048 ^ 2\)).

Perhaps the most unexpected results are the test cases for the DEAP framework, which are consistently slower than the EVAL baseline. In tree evaluation, we saw that domain calculation is slightly faster in DEAP than in our baseline. However, DEAP uses dynamic population sizes during evolution which might slow down the run. It is also worth mentioning that only the basic genetic operators and algorithms were used for DEAP. A more extensive experimentation with the evolutionary capabilities of this framework would most likely reveal a more optimal set of genetic operators and parameters that could prove faster than EVAL. Even so, that is not the aim of this work and so we shall compare TensorFlow timings against our baseline, which follows the same iterative principle.

In turn, EVAL proves to be slower than any of the TensorFlow approaches for all considered test cases, with an average verified speedup of almost 600 times over GPU\textunderscore TF\textunderscore EAGER for the \(512 ^ 2\) test case (262,144 points).

We can take the red line in Figure \ref{fig:mainfullrungraph} line as a visualization aid for approaches two orders of magnitude slower. For both iterative approaches, tests cases corresponding to higher problem sizes were not completed as they proved to be too time-consuming. Besides, based on previous results, performance margins would only maintain an increasing tendency.

%\captionsetup[figure]{font=small,skip=1pt}

\begin{figure}[hbt!]
    \centering
    \includegraphics[width=0.9\textwidth]{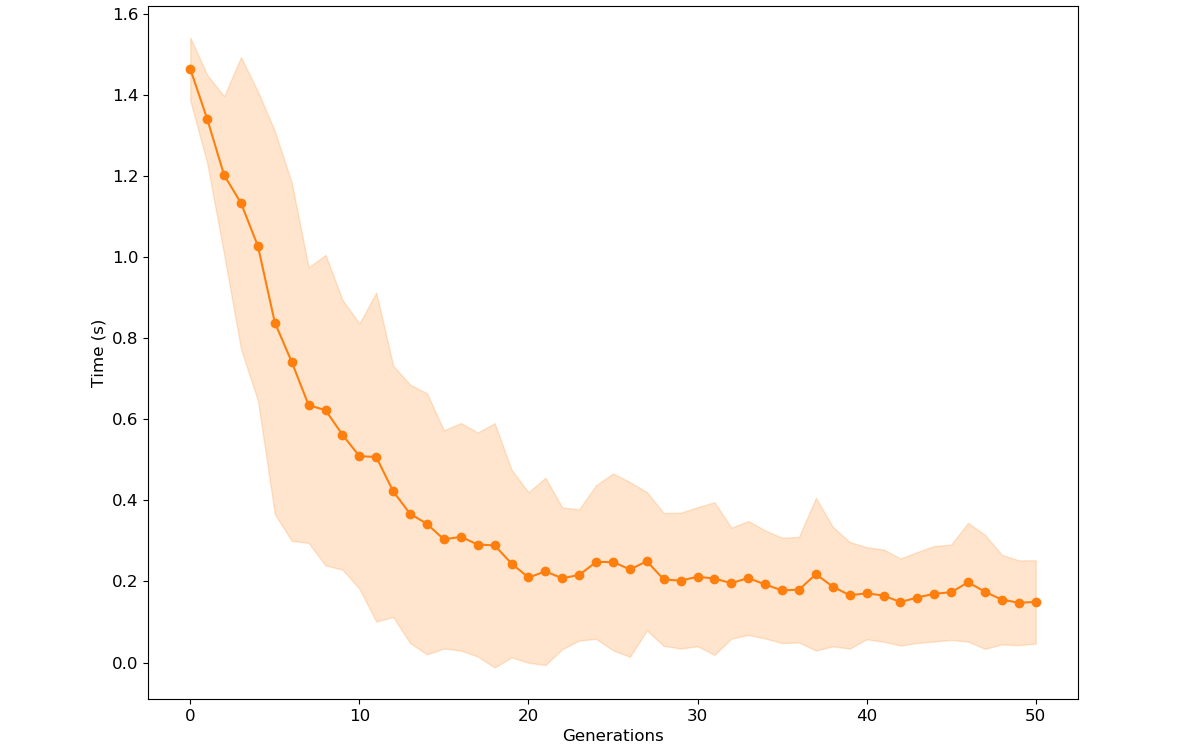}
    \caption{\small Average evaluation time (in seconds) cross generations for the 64 tensor side test case (4,096 points) with the GPU\textunderscore TF\textunderscore EAGER approach regarding the evolutionary experiment. Painted regions above and below represent one standard deviation from the average.}
    \label{fig:gensgovrum}
\end{figure}

\begin{figure}[hbt!]
    \centering
    \includegraphics[width=0.9\textwidth]{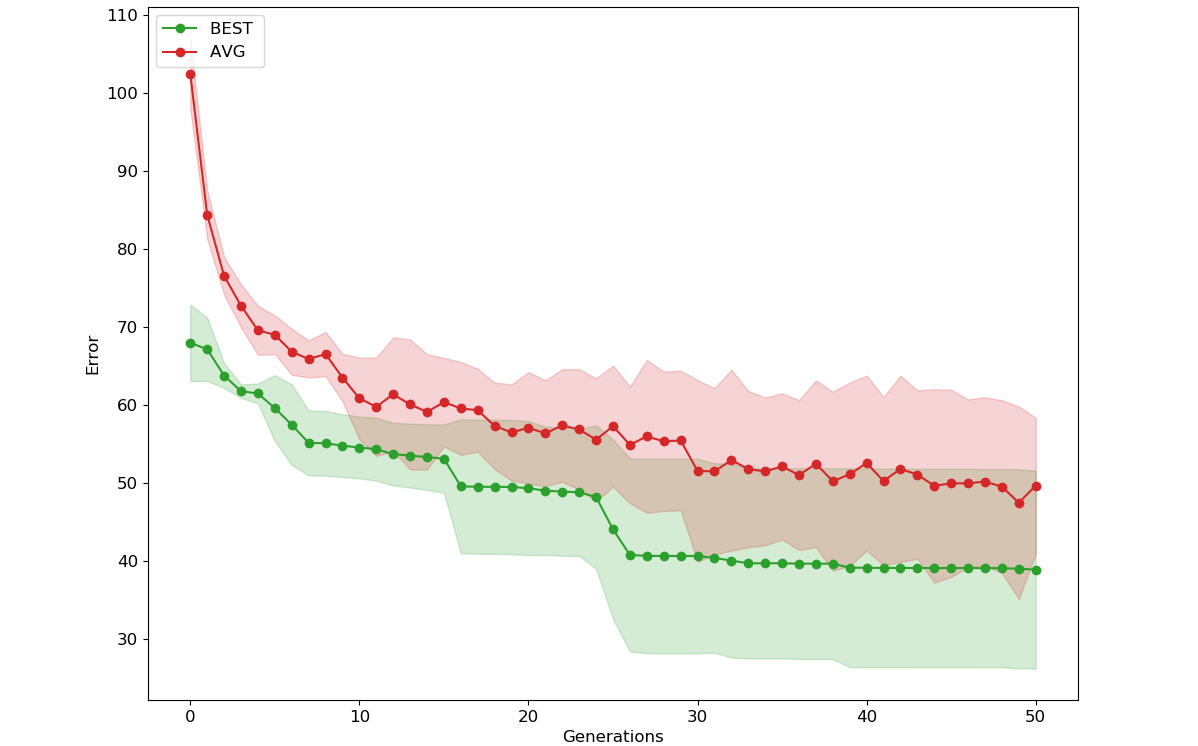}
    \caption{\small Average and minimum error values for the GPU\textunderscore TF\textunderscore EAGER approach regarding the \(2048^2\) test case for the evolutionary experiment. Painted regions above and below represent one standard deviation from the average.}
    \label{fig:evol}
\end{figure}

The results shown for the GPU in TensorFlow are fast, maybe even too fast. Indeed, with the evolutionary process thrown in the mix, it would be safe to assume that the speed up between iterative and vectorized approaches would shorten, even if marginally, as the genetic operators are run exclusively on CPU.
However, This seems to not be the case. 

In fact, speedups are higher when compared to tree evaluation experiments. Previous results with 512 tensor side (262,144) for TensorFlow GPU against EVAL regarding raw tree evaluation shows a speedup of almost 140 times, which is a far cry from the aforementioned nearly 600 times confirmed with evolution.
This can be explained by the caching of intermediate results that TensorFlow performs, leading to a pronounced decrease in evaluation time after the first few initial generations, as observed in Figure \ref{fig:gensgovrum}. These results further make the case for expression evolution with TensorFlow in eager mode.

Finally, to demonstrate TensorGP's capability for the evolution of large domains, in Figure \ref{fig:evol} we showcase fitness progression across generations for the test case with over 4 million points (\(2048^2\)).

%% file: 5-conclusion.tex
\section{Conclusions and Future Work} \label{sec:conclusions}

In this work, we propose different approaches to ease the computational burden of GP by taking advantage of its high potential for parallelism.
Namely, we investigate the advantages of applying data vectorization to the fitness evaluation phase using throughput-oriented architectures such as the GPU.
To accomplish this, we employed the TensorFlow numerical computation library written in Python, to develop a general-purpose GP engine capable of catering to our vectorization needs -- TensorGP.

Experimental results with our engine show that performance gains of up to 600 fold are attainable for the approximation of large evaluation domains in regards to the Pagie Polynomial function. Furthermore, we demonstrate the benefits of TensorFlow's eager execution model over graph execution for the caching of fitness results throughout generations.
Nevertheless, our test results for smaller domains seem to still make the case for more latency-oriented programming models such as the CPU. Therefore, modern-day GP seems to be best suited for heterogeneous computing frameworks like TensorFlow that are device-independent.

Upon completion of this work, some possibilities are to be considered for future endeavours.
We believe that the implementation of a preprocessing phase to simplify the mathematical expressions of individuals could greatly improve TensorGP's performance.
Additionally, we consider the elimination of overhead incurred by operator composition and protection mechanisms through the integration of device specific code with TensorFlow.
Moreover, with the incorporation of image specific operators in our engine, the exploration of evolutionary art is an appealing work path. 
Finally, the time comparison study amongst different approaches could be extended by including different GP frameworks, possibly under less strict evolutionary setups.

\section*{Acknowledgements}

This work is funded by national funds through the FCT - Foundation for Science and Technology, I.P., within the scope of the project CISUC - UID/CEC/00326/2020 and by European Social Fund, through the Regional Operational Program Centro 2020 and by the project grant DSAIPA/DS/0022/2018 (GADgET). We also thank the NVIDIA Corporation for the hardware granted to this research.

%% file: baeta.bbl
\begin{thebibliography}{10}
\providecommand{\url}[1]{\texttt{#1}}
\providecommand{\urlprefix}{URL }

\bibitem{9}
Giacobini, M., Tomassini, M., Vanneschi, L.: Limiting the number of fitness
  cases in genetic programming using statistics. In: International Conference
  on Parallel Problem Solving from Nature. pp. 371--380. Springer (2002)

\bibitem{poli2008field}
Poli, R., Langdon, W.B., McPhee, N.F., Koza, J.R.: A field guide to genetic
  programming. Lulu. com (2008)

\bibitem{10}
Andre, D., Koza, J.R.: Parallel genetic programming: A scalable implementation
  using the transputer network architecture. In: Advances in genetic
  programming. pp. 317--337. MIT Press (1996)

\bibitem{16}
Moore, G.E., et~al.: Cramming more components onto integrated circuits (1965)

\bibitem{arenas2012gpu}
Arenas, M., Romero, G., Mora, A., Castillo, P., Merelo, J.: Gpu parallel
  computation in bioinspired algorithms: a review. In: Advances in Intelligent
  Modelling and Simulation, pp. 113--134. Springer (2012)

\bibitem{24}
Staats, K., Pantridge, E., Cavaglia, M., Milovanov, I., Aniyan, A.: Tensorflow
  enabled genetic programming. In: Proceedings of the Genetic and Evolutionary
  Computation Conference Companion. pp. 1872--1879. ACM (2017)

\bibitem{5}
Keijzer, M.: Efficiently representing populations in genetic programming. In:
  Advances in genetic programming. pp. 259--278. MIT Press (1996)

\bibitem{4}
Handley, S.: On the use of a directed acyclic graph to represent a population
  of computer programs. In: Proceedings of the First IEEE Conference on
  Evolutionary Computation. IEEE World Congress on Computational Intelligence.
  pp. 154--159. IEEE (1994)

\bibitem{7}
Keijzer, M.: Alternatives in subtree caching for genetic programming. In:
  European Conference on Genetic Programming. pp. 328--337. Springer (2004)

\bibitem{8}
Wong, P., Zhang, M.: Scheme: Caching subtrees in genetic programming. In: 2008
  IEEE Congress on Evolutionary Computation (IEEE World Congress on
  Computational Intelligence). pp. 2678--2685. IEEE (2008)

\bibitem{machado1999speeding}
Machado, P., Cardoso, A.: Speeding up genetic programming. In: Procs. 2nd Int.
  Symp. AI and Adaptive Systems, CIMAF. vol.~99, pp. 217--222 (1999)

\bibitem{26}
Chitty, D.M.: Fast parallel genetic programming: multi-core cpu versus
  many-core gpu. Soft Computing  16(10),  1795--1814 (2012)

\bibitem{111}
Burlacu, B., Kronberger, G., Kommenda, M.: Operon c++ an efficient genetic
  programming framework for symbolic regression. In: Proceedings of the 2020
  Genetic and Evolutionary Computation Conference Companion. pp. 1562--1570
  (2020)

\bibitem{de2020mimd}
de~Melo, V.V., Fazenda, {\'A}.L., Sotto, L.F.D.P., Iacca, G.: A mimd
  interpreter for genetic programming. In: International Conference on the
  Applications of Evolutionary Computation (Part of EvoStar). pp. 645--658.
  Springer (2020)

\bibitem{19}
Cano, A., Zafra, A., Ventura, S.: Speeding up the evaluation phase of gp
  classification algorithms on gpus. Soft Computing  16(2),  187--202 (2012)

\bibitem{14}
Chitty, D.M.: A data parallel approach to genetic programming using
  programmable graphics hardware. In: Proceedings of the 9th annual conference
  on Genetic and evolutionary computation. pp. 1566--1573. ACM (2007)

\bibitem{13}
Cano, A., Ventura, S.: Gpu-parallel subtree interpreter for genetic
  programming. In: Proceedings of the 2014 Annual Conference on Genetic and
  Evolutionary Computation. pp. 887--894. ACM (2014)

\bibitem{20}
Augusto, D.A., Barbosa, H.J.: Accelerated parallel genetic programming tree
  evaluation with opencl. Journal of Parallel and Distributed Computing  73(1),
   86--100 (2013)

\bibitem{25}
Koza, J.R., Bennett, F., Hutchings, J.L., Bade, S.L., Keane, M.A., Andre, D.:
  Evolving sorting networks using genetic programming and the rapidly
  reconfigurable xilinx 6216 field-programmable gate array. In: Conference
  Record of the Thirty-First Asilomar Conference on Signals, Systems and
  Computers (Cat. No. 97CB36136). vol.~1, pp. 404--410. IEEE (1997)

\bibitem{tensorflow}
Abadi, M., Barham, P., Chen, J., Chen, Z., Davis, A., Dean, J., Devin, M.,
  Ghemawat, S., Irving, G., Isard, M., et~al.: Tensorflow: A system for
  large-scale machine learning. In: 12th USENIX Symposium on Operating Systems
  Design and Implementation (OSDI 16). pp. 265--283 (2016)

\bibitem{kai1}
Cavaglia, M., Staats, K., Gill, T.: Finding the origin of noise transients in
  ligo data with machine learning. arXiv preprint arXiv:1812.05225  (2018)

\bibitem{kai2}
Fu, X., Ren, X., Mengshoel, O.J., Wu, X.: Stochastic optimization for market
  return prediction using financial knowledge graph. In: 2018 IEEE
  International Conference on Big Knowledge (ICBK). pp. 25--32. IEEE (2018)

\bibitem{kai3}
Matousek, R., Hulka, T., Dobrovsky, L., Kudela, J.: Sum epsilon-tube error
  fitness function design for gp symbolic regression: Preliminary study. In:
  2019 International Conference on Control, Artificial Intelligence, Robotics
  \& Optimization (ICCAIRO). pp. 78--83. IEEE (2019)

\bibitem{wolfram}
Rowland, T., Weisstein, E.W.: Tensor. {From MathWorld---A Wolfram Web
  Resource}, \url{http://mathworld.wolfram.com/Tensor.html}

\bibitem{agrawal2019tensorflow}
Agrawal, A., Modi, A.N., Passos, A., Lavoie, A., Agarwal, A., Shankar, A.,
  Ganichev, I., Levenberg, J., Hong, M., Monga, R., et~al.: Tensorflow eager: A
  multi-stage, python-embedded dsl for machine learning. arXiv preprint
  arXiv:1903.01855  (2019)

\bibitem{pagie1997evolutionary}
Pagie, L., Hogeweg, P.: Evolutionary consequences of coevolving targets.
  Evolutionary computation  5(4),  401--418 (1997)

\bibitem{deap}
Fortin, F.A., De~Rainville, F.M., Gardner, M.A.G., Parizeau, M., Gagn{\'e}, C.:
  Deap: Evolutionary algorithms made easy. The Journal of Machine Learning
  Research  13(1),  2171--2175 (2012)

\end{thebibliography}
